\documentclass[11pt]{article}

\usepackage[final]{acl}

\usepackage{times}
\usepackage{latexsym}
\usepackage[T1]{fontenc}
\usepackage[utf8]{inputenc}
\usepackage{microtype}
\usepackage{graphicx}
\usepackage{inconsolata}
\usepackage{amsmath}
\usepackage{amssymb}
\usepackage{booktabs}
\usepackage{multirow}
\usepackage{array}
\usepackage{xspace}
\usepackage{url}
\usepackage[table]{xcolor}
\usepackage{makecell}
\usepackage{subcaption}
\newcolumntype{L}[1]{>{\raggedright\arraybackslash}p{#1}}

\newcommand{\tqs}{\textsc{tqs}\xspace}
\newcommand{\OEC}{\textsc{oec}\xspace}
\newcommand{\IAC}{\textsc{iac}\xspace}
\newcommand{\PSN}{\textsc{psn}\xspace}
\newcommand{\TTE}{\textsc{tte}\xspace}
\newcommand{\SCP}{\textsc{scp}\xspace}
\newcommand{\HCR}{\textsc{hcr}\xspace}
\newcommand{\PED}{\textsc{ped}\xspace}
\newcommand{\DPI}{\textsc{dpi}\xspace}
\newcommand{\swev}{SWE-Bench Verified\xspace}
\newcommand{\openhands}{OpenHands\xspace}
\newcommand{\coderforge}{CoderForge-Preview\xspace}
\newcommand{\ph}[1]{\textbf{\textcolor{black}{#1}}}

\title{Quality, Not Just Outcome: A Structural Scoring Framework for Selecting Coding-Agent Trajectories for Supervised Fine-Tuning}

\author{
\textbf{Sida Li}\textsuperscript{1} \quad
\textbf{Yuanxin Liu}\textsuperscript{1} \quad
\textbf{Shicheng Li}\textsuperscript{1} \quad
\textbf{Linli Yao}\textsuperscript{1} \quad
\textbf{Xu Sun}\textsuperscript{1} \\
\textbf{Haoli Bai}\textsuperscript{2} \quad
\textbf{Xiao-Hui Li}\textsuperscript{2}
\\[1mm]
\textsuperscript{1}National Key Laboratory for Multimedia Information Processing, Peking University \\
\textsuperscript{2}Huawei Technologies Co., Ltd. \\
\small{
\texttt{\{lisida,liuyuanxin,linliyao\}@stu.pku.edu.cn} \quad
\texttt{\{lisc99,xusun\}@pku.edu.cn}
}\\
\small{
\texttt{\{baihaoli,lixiaohui33\}@huawei.com}
}
}

\begin{document}
\maketitle

\begin{abstract}
Open coding agents are commonly trained through supervised fine-tuning on interaction trajectories whose final patches pass the benchmark test suite. However, this outcome-level filtering criterion ignores process quality: trajectories that eventually succeed may still contain inefficient interaction patterns such as repetitive file inspection, unstable tool usage, or delayed modification behavior, all of which become part of the imitation target. We propose \tqs\ (Trajectory Quality Score), a structural scoring framework for coding-agent trajectories that operates directly on trajectory structure, requires no LLM judge, and runs in linear time. \tqs\ characterizes trajectory quality through structural signals organized around structured convergence: information focus, intent--action alignment, and policy dynamics. Using 155{,}144 successful trajectories from a public open-license corpus, we rank trajectories by \tqs\ and fine-tune coding agents on filtered subsets. Models trained on top-ranked trajectories consistently outperform those trained on random or bottom-ranked subsets on \swev. Our results suggest that process-level trajectory quality is an important factor in coding-agent supervision and that structural trajectory filtering provides an effective and scalable alternative to LLM-based data selection. Project page: \url{https://github.com/DuHeng0128/TQS.git}.
\end{abstract}

\section{Introduction}

Coding agents that interact with sandboxed environments to inspect repositories, execute tests, and edit files have recently achieved strong performance on \swev-style software engineering tasks~\citep{jimenez2024swebench,openai2024verified}. Beyond advances in foundation models and inference-time scaffolding, an increasingly important ingredient in these systems is trajectory supervision: training on multi-turn agent demonstrations consisting of reasoning traces, tool calls, tool observations, and final patches~\citep{pan2025swegym,ma2024lingma,ariyak2026coderforge}. For medium-scale open-weight models, a common and effective practice is to collect such trajectories from more capable teacher agents and retain those whose final patches pass the benchmark test suite.

However, outcome-level correctness does not necessarily imply process-level quality. Among successful trajectories, behavioral patterns vary substantially. Some trajectories localize the bug quickly, maintain coherent search scope, and transition efficiently from exploration to modification. Others repeatedly inspect the same files, interrupt stalled shell commands, or postpone meaningful edits until late in the interaction budget, yet still eventually reach a passing patch through extended trial-and-error. These differences matter because trajectory-level supervised fine-tuning exposes the student to the full interaction process, including reasoning traces, tool use, and the final patch~\citep{kamoi2024when}. Inefficient or unstable interaction patterns therefore become part of the training signal.

\begin{figure*}[t]
\centering
\includegraphics[width=\textwidth]{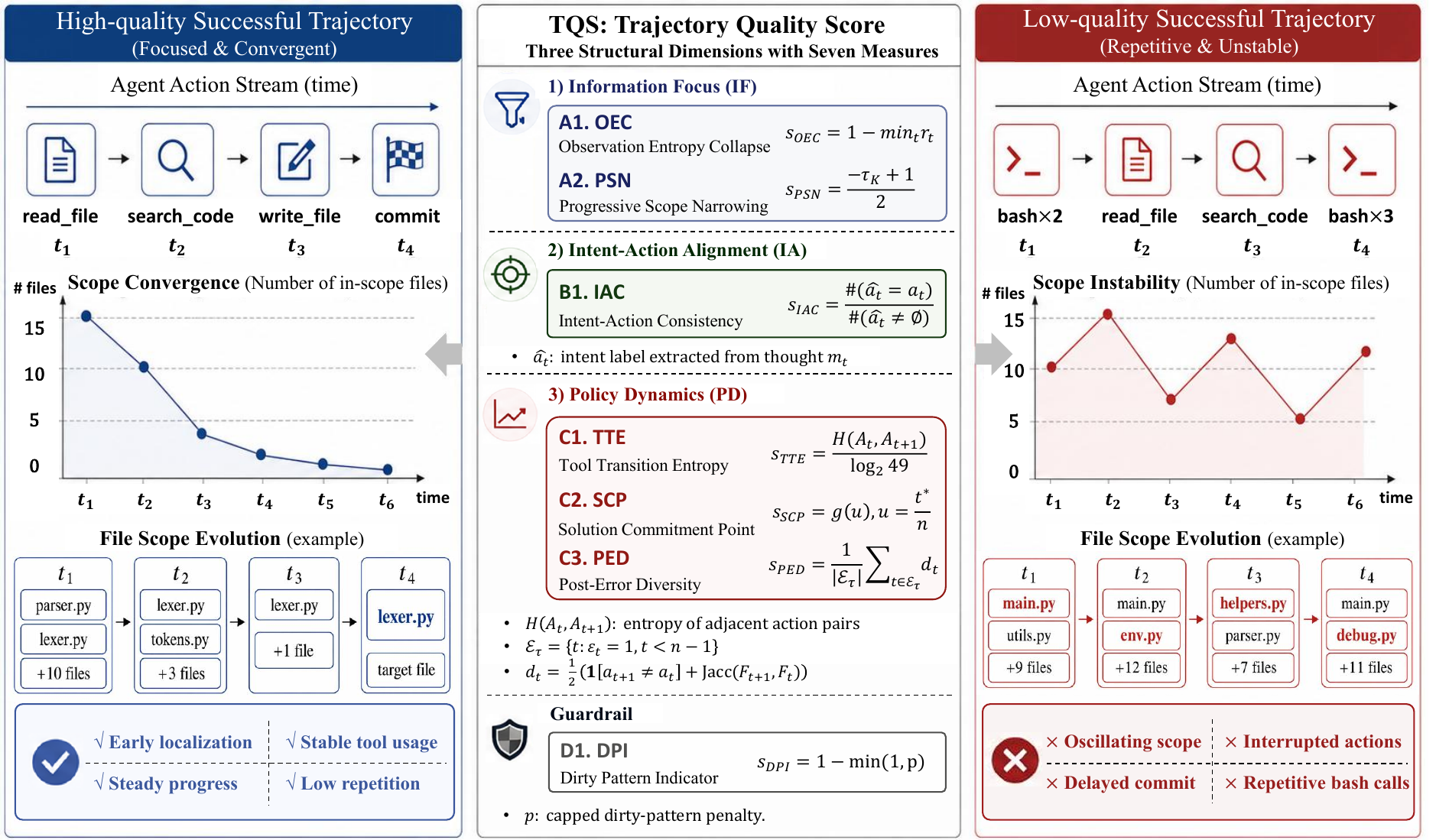}
\caption{\textbf{Overview of \tqs.} Contrasting high-quality (focused and convergent) with low-quality (repetitive and unstable) successful trajectories, TQS evaluates trajectory quality along three structural dimensions: Information Focus (OEC, PSN), Intent–Action Alignment (IAC), and Policy Dynamics (TTE, SCP, PED), with a Dirty Pattern Indicator (DPI) guardrail. These measures are aggregated into a weighted score in \([0,1]\).}
\label{fig:pipeline}
\end{figure*}

The instruction-tuning literature has shown that curated subsets can outperform larger, noisier training corpora~\citep{zhou2023lima,chen2024alpagasus,liu2024deita,li2024cherry,xia2024less}. Coding-agent trajectories pose a different selection problem. They are long interaction histories whose ordering records exploration, editing, testing, and recovery. Perplexity-style scores are confounded by length and tokenization, while LLM-based judges are costly at the scale of hundreds of thousands of trajectories and billions of tokens. Furthermore, these approaches are not designed to characterize how effectively an agent organizes its search process over time.

In this work, we propose \tqs\ (Trajectory Quality Score), a structural scoring framework for OpenHands-style coding-agent trajectories. The framework is built around \emph{structured convergence}: as an agent approaches a solution, a strong trajectory becomes increasingly organized around the relevant fix. This organization appears at three levels. The agent should focus its observations and file operations on task-relevant evidence, keep its stated intent aligned with its tool actions, and move from broad exploration toward productive edit, test, and recovery behavior. We operationalize these levels through seven trajectory-level measures, each computable in linear time from event logs alone. Outcome labels are used only to validate candidate structural signals during metric development; once the components are fixed, \tqs\ ranks successful trajectories without inspecting the final patch, held-out test result, or downstream SFT performance.

Empirically, we score 155,144 successful trajectories from \coderforge~\citep{ariyak2026coderforge} and construct matched top-ranked, bottom-ranked, and random subsets from the eligible fixed-budget SFT pool under task- and repository-level diversity constraints. We LoRA-fine-tune the same Qwen3-Coder-30B-A3B-Instruct model~\citep{ariyak2026coderforge,yang2025qwen3,hu2022lora} on each subset using identical hyperparameters. Models trained on top-ranked trajectories consistently outperform those trained on random or bottom-ranked subsets on \swev~\citep{openai2024verified} under the \openhands framework~\citep{wang2025openhands}. These results suggest that trajectory-level process structure is an important and previously underexplored dimension of coding-agent supervision.

Our contributions are threefold:
(i) we formulate coding-agent trajectory quality through structured convergence and operationalize it with seven efficient trajectory-level measures;
(ii) we introduce a scalable framework for trajectory scoring and filtering without relying on LLM judges or external reward models; and
(iii) we present an end-to-end open pipeline for trajectory scoring, filtering, fine-tuning, and evaluation, demonstrating that quality-aware trajectory selection improves downstream coding-agent performance.

\section{Related Work}

\paragraph{Coding agents and trajectory supervision.}
Recent coding agents interact with repositories through tool-augmented environments that support file inspection, shell execution, testing, and patch generation. Frameworks such as SWE-agent~\citep{yang2024sweagent} and \openhands~\citep{wang2025openhands} standardized the agent-computer interaction pipeline and enabled large-scale collection of executable agent trajectories. Building on these infrastructures, datasets such as SWE-Gym~\citep{pan2025swegym}, Lingma SWE-GPT~\citep{ma2024lingma}, and \coderforge~\citep{ariyak2026coderforge} progressively scaled teacher-generated trajectory supervision for open-weight coding agents. Recent work has also explored trajectory synthesis and reward-guided curation for software and GUI agents~\citep{sun2025genesis}. While reinforcement-learning approaches for software engineering agents~\citep{wei2025swerl} further improve exploration and decision-making, they typically still rely on strong supervised initialization, making trajectory quality important for both SFT and RL-based pipelines.

\paragraph{Data selection for supervised fine-tuning.}
Data-selection methods for supervised fine-tuning usually use an external judge or a training-derived importance signal. AlpaGasus~\citep{chen2024alpagasus} and DEITA~\citep{liu2024deita} ask stronger LLMs to score samples. LESS~\citep{xia2024less}, Cherry-LLM~\citep{li2024cherry}, and Superfiltering~\citep{li2024superfiltering} estimate utility through gradient influence or model-internal signals; related methods have also been applied to code instruction tuning~\citep{wang2024xcoder}. These selectors were developed for short, single-turn samples. Applying them to an agent trajectory would require redefining their sample-level signals over a long interaction history. \tqs instead uses the temporal structure already present in the event log.

\paragraph{Process-level evaluation and trajectory quality.}
Recent work has increasingly emphasized evaluating intermediate reasoning and interaction processes rather than final outcomes alone. Prior studies on chain-of-thought faithfulness analyze whether intermediate reasoning traces are consistent with the eventual answer~\citep{lanham2023faithfulness,bentham2024unfaithfulness}. AgentDiagnose~\citep{ou2025agentdiagnose} applies an LLM rubric to browser-agent trajectories and shows that process-aware trajectory selection can improve training efficiency. Beyond this, several methods use entropy or behavioral diversity to analyze and optimize agent policies~\citep{chen2026beyondquantity,li2026tepo,dong2025arpo,yu2025entropo}. Recent work has also explored process supervision and trajectory curation for software-engineering agents using privileged patch-level information~\citep{ma2026patches}.

\section{Trajectory Quality Score}
\label{sec:tqs}

\subsection{Preliminaries and Notation}

A trajectory $\tau = (s_0, s_1, \dots, s_{n-1})$ is a sequence of $n$ steps. Each step is represented as
\[
s_t = (a_t, m_t, o_t, F_t, \varepsilon_t),
\]
where $a_t \in \mathcal{A}$ is an action type (one of \texttt{think}, \texttt{read\_file}, \texttt{write\_file}, \texttt{execute\_bash}, \texttt{search}, \texttt{submit}, \texttt{other}); $m_t$ is the free-text thought that precedes the action; $o_t$ is the tool observation returned after the action; $F_t \subseteq \mathcal{F}$ is the set of file paths touched by the action; and $\varepsilon_t \in \{0,1\}$ indicates whether $o_t$ matches a predefined error pattern. This tuple is recoverable from any \openhands-style event log without model inference.
 
\subsection{Design Intuition and Score Construction}
\label{sec:frame}
 
A coding-agent trajectory is a search process over observations, files, actions, and intermediate hypotheses. We view a high-quality trajectory as one that exhibits \emph{structured convergence}: over the course of the interaction, the search becomes organized around the emerging fix. \tqs\ relies on event-log evidence that the agent is narrowing its search toward a task-relevant region. At minimum, it provides a low-cost way to filter out clearly undesirable trajectory patterns.
 
Structured convergence has three observable levels. \textbf{(A) Information focus} asks whether the agent narrows the relevant evidence through its observations and file operations. \textbf{(B) Intent--action alignment} asks whether the stated thought before a tool call points in the same direction as the action that follows. \textbf{(C) Policy dynamics} asks whether the interaction policy develops a productive rhythm of exploration, execution, and error recovery. Some failures are categorical rather than gradual; the trajectories associated with them are inefficient. A run may trigger an early-stop pattern, never make a substantive edit, or repeat the same failing command. \DPI therefore sits outside the three-dimensional organization as a guardrail for these cases.
 
The axes motivate candidate measures, and the selection protocol decides which ones enter \tqs. Each component is computable from the step tuple in $O(n)$ CPU time without auxiliary LLM judging, so the scorer can process event logs spanning roughly 6B tokens. The checks in \S\ref{sec:validity} retain measures that discriminate above chance, have limited correlation with step count, and preserve behavioral diversity during subset selection.

\subsection{Components and Aggregation}
\label{sec:components}

\tqs uses seven components, each normalized to $[0,1]$ with larger values indicating cleaner structure. A component is left undefined when its required event is absent; Eq.~\ref{eq:tqs} handles these cases during aggregation.

\paragraph{(A1) Observation Entropy Collapse (\OEC).}
For each observation $o_t$, we compute character-level Shannon entropy~\citep{shannon1948}. Let $\widetilde H_t$ be its three-step moving average, $\overline H_{\mathrm{open}}$ the mean entropy of the first $\min(5,n)$ observations, and $r_t=\widetilde H_t/\overline H_{\mathrm{open}}$. Then
\[
\begin{gathered}
H(o_t)=-\sum_c p_t(c)\log_2 p_t(c),\\
s_{\OEC}=\mathrm{clamp}\!\left(1-\min_t r_t,0,1\right).
\end{gathered}
\]
Lower relative entropy indicates that observations have concentrated around a smaller body of evidence. Since an unproductive loop can produce the same pattern, \OEC is used with \TTE and \DPI.

\paragraph{(A2) Progressive Scope Narrowing (\PSN).}
We count the distinct files touched in a five-step window and compute Kendall's $\tau_K$~\citep{kendall1938} between this scope and the step index over the second half of the trajectory:
\[
\begin{gathered}
\mathrm{scope}_t=\left|\bigcup_{k=\max(0,t-4)}^t F_k\right|,\\
s_{\PSN}=\mathrm{clamp}\!\left(\frac{1-\tau_K}{2},0,1\right).
\end{gathered}
\]
A shrinking active file set gives negative $\tau_K$ and therefore a higher score, matching the localization pattern used by coding agents~\citep{xia2024agentless,yu2025orcaloca}.

\paragraph{(B1) Intent--Action Consistency (\IAC).}
A fixed lexicon-based parser maps the thought preceding each
non-\texttt{think} action to a coarse intended action $\hat a_t$: reading, editing, execution/testing, or search. For the extractable steps
$\mathcal I=\{t:\hat a_t\text{ is defined}\}$,
\[
s_{\IAC}=\frac{1}{|\mathcal I|}
\sum_{t\in\mathcal I}\mathbf 1[\hat a_t=a_t].
\]
\IAC is undefined when $\mathcal I=\varnothing$. It captures coarse thought--action consistency characteristic of ReAct-style
interaction~\citep{yao2023react}, without assessing semantic correctness.

\paragraph{(C1) Tool Transition Entropy (\TTE).}
Let $p(i\!\to\!j)$ be the empirical frequency of adjacent transition $(i,j)$ in the seven-action alphabet. We normalize its entropy by the maximum over 49 transition types:
\[
s_{\TTE}=\frac{-\sum_{i,j}p(i\!\to\!j)\log_2 p(i\!\to\!j)}{\log_2 49}.
\]
\TTE penalizes long single-tool loops. Focus and guardrail measures provide the context needed to distinguish useful variation from erratic switching.

\paragraph{(C2) Solution Commitment Point (\SCP).}
A substantive edit is a \texttt{write\_file} action whose observation contains no error pattern. Let $t^*$ be the first such edit, $u=t^*/n$, $d(u)=\max(0,0.2-u,u-0.5)$, and $\sigma=0.2$. We set
\[
s_{\SCP}=\left\{
\begin{array}{@{}l@{}r@{}}
0, & \text{no substantive edit};\\
\exp\!\left(-d(u)^2/(2\sigma^2)\right), & \text{otherwise}.
\end{array}
\right.
\]
The score is one for $u\in[0.2,0.5]$ and decays for earlier or later commitment. Because $u$ is relative time, \SCP does not prefer a particular absolute trajectory length.

\paragraph{(C3) Post-Error Diversity (\PED).}
Let $\mathcal E_\tau=\{t:\varepsilon_t=1,t<n-1\}$ be the error steps followed by an action. We combine a change in action type with the Jaccard distance between successive file sets:
\[
\begin{gathered}
\mathrm{Jacc}(F_{t+1},F_t)
=1-\frac{|F_{t+1}\cap F_t|}{|F_{t+1}\cup F_t|},\\
d_t=\tfrac12\!\left(
\mathbf 1[a_{t+1}\ne a_t]
+\mathrm{Jacc}(F_{t+1},F_t)
\right),\\
s_{\PED}
=\frac{1}{|\mathcal E_\tau|}
\sum_{t\in\mathcal E_\tau} d_t,
\end{gathered}
\]
\PED is undefined when $\mathcal E_\tau$ is empty. Repeating the same action on the same files gives zero; changing the tool, file scope, or both raises the score~\citep{shinn2023reflexion,kamoi2024when}.

\paragraph{(D1) Dirty Pattern Indicator (\DPI).}
\DPI captures categorical failures that may survive the continuous measures: early stopping, no substantive edit, repeated shell-command loops, and repeated errors. With repeated-loop fraction \emph{lp} and error-repetition rate \emph{er},
\[
\begin{gathered}
p=0.4\,\mathbf 1[\mathrm{early}]+0.4\,\mathbf 1[\mathrm{no\ edit}]\\
{}+\min(0.3,3\,\mathrm{lp})+\min(0.3,0.6\,\mathrm{er}),\\
s_{\DPI}=1-\min(1,p).
\end{gathered}
\]
A trajectory with no flagged pattern receives full guardrail credit; each detected pattern reduces the contribution of \DPI.

\paragraph{Aggregation.}

We use fixed, hand-set weights throughout: $w_{\OEC}=0.30$, $w_{\IAC}=0.25$, $w_{\PSN}=w_{\TTE}=w_{\PED}=w_{\DPI}=0.10$, and $w_{\SCP}=0.05$. The resulting totals are 0.40 for information focus, 0.25 for intent--action alignment, 0.25 for policy dynamics, and 0.10 for the guardrail. Perturbing $w_{\OEC}$, $w_{\IAC}$, or $w_{\DPI}$ one at a time retains 93.1--99.6\% of the Top-1{,}024 selection (Appendix~\ref{app:weights}).

The final \tqs\ is a renormalized weighted mean over the components defined for a trajectory:
\begin{equation}
\tqs(\tau) \;=\; \frac{\sum_{c \in \mathcal{C}^*(\tau)} w_c \cdot s_c(\tau)}{\sum_{c \in \mathcal{C}^*(\tau)} w_c},
\label{eq:tqs}
\end{equation}
where $\mathcal{C}^*(\tau)$ is the set of components defined on $\tau$. Trajectories with $n<3$ or component coverage below $4/7$ are excluded. Outcome labels are used only for component admission during metric development (as described in \S\ref{sec:validity}); after the components and weights are fixed, \tqs\ ranks successful trajectories using event-log structure alone.
 
\subsection{Discriminant Validity and External Sanity Checks}
\label{sec:validity}
 
We use task outcome as a weak admission signal for candidate components. Ineffective process patterns often make failure more likely, so a useful component should tend to rank a passing trajectory above a failing trajectory from the same task. This comparison filters signals with the wrong direction or strong confounds. Once the components are fixed, \tqs ranks only task-successful trajectories for training, using event-log structure alone and no further patch-level supervision.

Concretely, we pair each passing trajectory with a failing trajectory from the same task. For each candidate measure, we compute two quantities. The first is concordance: the rate at which the measure assigns a higher score to the passing trajectory than to its paired failing trajectory, with 0.5 corresponding to chance. The second is the absolute Spearman correlation $|\rho|$ with trajectory length, which serves as a simple check against length-driven proxies. A measure is admitted only if its concordance exceeds 0.5 and $|\rho| < 0.25$.

\begin{figure}[t]
\centering

\small
\setlength{\tabcolsep}{2.8pt}
\renewcommand{\arraystretch}{1.05}

\begin{tabular}{@{}l c c l@{}}
\toprule
\textbf{Measure} & \textbf{Conc.} & \textbf{$|\rho|$} & \textbf{Outcome} \\
\midrule
OEC & 0.640 & 0.103 & \checkmark kept \\
IAC & 0.612 & 0.087 & \checkmark kept \\
PSN & 0.585 & 0.142 & \checkmark kept \\
SCP & 0.598 & 0.058 & \checkmark kept \\
TTE & 0.563 & 0.218 & \checkmark kept \\
PED & 0.541 & 0.034 & \checkmark kept \\
DPI & 0.591 & 0.071 & \checkmark guardrail \\
\midrule
TQS Composite & \textbf{0.732} & 0.092 & --- \\
\midrule
HCR & 0.527 & 0.014 & \makecell[l]{$\times$ diversity} \\
BUS/TGCR/DCS-f. & $\le 0.479$ & $<0.25$ & \makecell[l]{$\times$ conc.} \\
DCS-Jacc. & --- & 0.729 & \makecell[l]{$\times$ length} \\
\bottomrule
\end{tabular}

\captionof{table}{
Discriminant validity on paired success/failure trajectories from \coderforge. \tqs's composite concordance exceeds every individual component. Dropped measures reflect the role of admission protocol: \HCR is rejected for behavioral mode collapse, BUS/TGCR/DCS-file for low concordance, and DCS-Jaccard for length correlation.
}
\label{tab:validity}

\includegraphics[width=\linewidth]{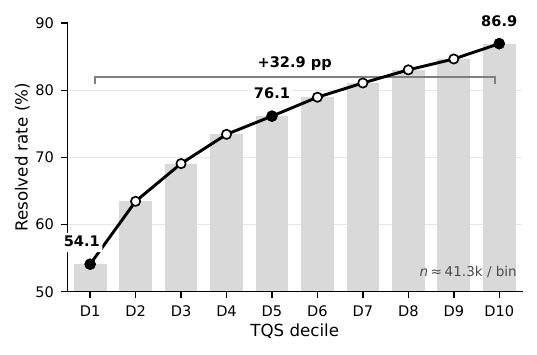}
\captionof{figure}{
External validity of \tqs on the released scoring records. We bin 413{,}278 records into ten equal-frequency deciles by \tqs. These records include the source shards and the release's reward-1 convenience shards. The resolved rate increases monotonically from 54.1\% in D1 to 86.9\% in D10.
}
\label{fig:tqs-deciles}
\end{figure}

Table~\ref{tab:validity} summarizes the results. The seven retained components pass both criteria, and the composite \tqs\ reaches a concordance of 0.732, above every individual component. The same criteria exclude several candidates: BUS, TGCR, and DCS-file fail the concordance criterion, while DCS-Jaccard is dominated by length correlation ($|\rho|=0.729$). \HCR\ passes the numerical thresholds, but including it caused the Top-1K set to concentrate 99.9\% of its mass on a single binary behavior pattern. This mode collapse is invisible to pairwise concordance but matters for SFT, where trajectories should preserve behavioral diversity. We therefore add behavioral-diversity preservation as a third admission criterion and exclude \HCR\ under it. Additional details on the rejected measures are provided in Appendix~\ref{app:failed}.

\begin{figure*}[t]
\centering
\includegraphics[width=\textwidth]{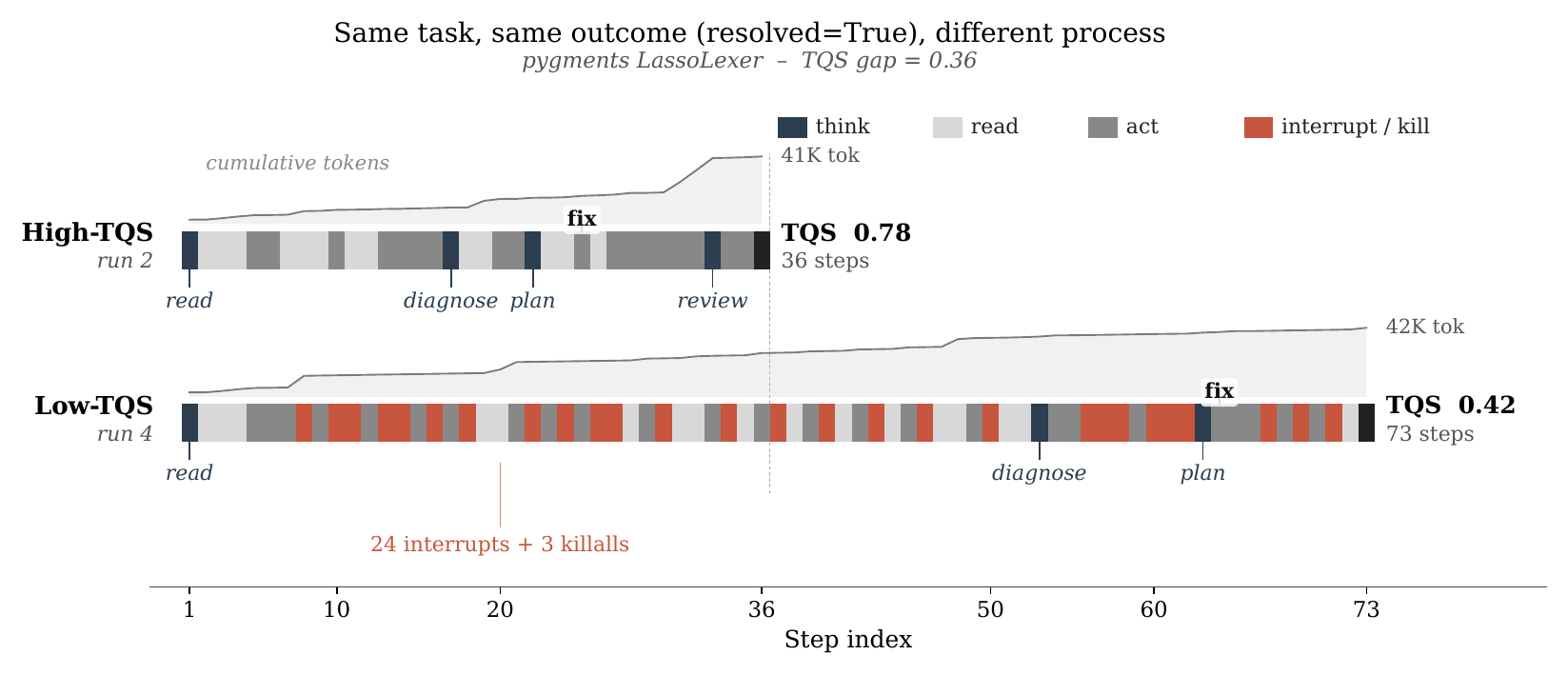}
\caption{Two passing trajectories on the same \texttt{pygments} \texttt{LassoLexer} task. The high-\tqs\ run reaches the fix in 36 steps. The low-\tqs\ run takes 73 steps and includes many interrupts and forced process terminations. Their final outcomes are equal, and their \tqs values differ by 0.36.}
\label{fig:case_study}
\end{figure*}

\paragraph{External outcome check.}
We also compare the finalized score with the test harness's \texttt{resolved} label, which is absent from the scoring function. Figure~\ref{fig:tqs-deciles} shows a monotonic rise in resolved rate across the ten \tqs\ deciles, from 54.1\% in D1 to 86.9\% in D10. The 32.9-point spread has no inversion at a decile boundary. Within a fixed step-count band, pair-AUC is 1.000 for the composite and above 0.96 for \OEC, \IAC, and \PED\ (Appendix~\ref{app:lengthauc}).
 

\section{Experimental Setup}

\paragraph{Qualitative motivation.}

Figure~\ref{fig:case_study} shows two passing \openhands runs on the same \texttt{pygments} \texttt{LassoLexer} task. Both resolve the task. The high-\tqs trajectory reaches the fix in 36 steps through a compact read--diagnose--plan--review pattern; the low-\tqs trajectory takes 73 steps and contains many interrupt and kill actions. The largest component gaps occur in observation focus (\OEC: 0.738 vs.\ 0.312), commitment timing (\SCP: 0.990 vs.\ 0.528), and post-error adaptation (\PED: 1.000 vs.\ 0.000), revealing substantial behavioral differences that the final outcome alone does not capture. Appendix~\ref{app:worked-example} reports all seven component scores.

\paragraph{Data and subset construction.}
\begin{table}[h]
\centering
\small
\setlength{\tabcolsep}{4pt}
\begin{tabular}{lccc}
\toprule
\textbf{Subset} & \textbf{Top} & \textbf{Bottom} & \textbf{Random} \\
\midrule
\# trajectories        & 1{,}024 & 1{,}024 & 1{,}024 \\
Mean \tqs              & 0.714 & 0.454 & 0.589 \\
Mean steps             & 34.2 & 53.0 & 41.0 \\
Mean tokens (K)        & 28.1 & 26.3 & 26.9 \\
Tasks / repos          & 847/209 & 847/204 & 825/217 \\
Max per task           & 30 & 30 & 59 \\
Source mix (\%)        & 61/20/18 & 61/20/18 & 61/20/18 \\
Dirty flag (\%)        & 1.0 & 64.3 & 13.6 \\
\bottomrule
\end{tabular}
\caption{Training subsets. Top and Bottom have zero \tqs overlap (Top.min $=0.706 >$ Bottom.max $=0.470$). Source mix is Smith/Reb./Gym.}
\label{tab:subsetstats}
\end{table}
We score \coderforge~\citep{ariyak2026coderforge}, an open-license corpus of \ph{258{,}134} unique OpenHands-format trajectories produced by a Qwen3-Coder-480B teacher on \ph{51{,}201} tasks from R2E-Gym~\citep{jain2025r2egym}, SWE-Smith~\citep{yang2025swesmith}, and SWE-Rebench~\citep{badertdinov2025swerebench}. The \ph{155{,}144} passing trajectories form the candidate pool and were deduplicated by \texttt{trajectory\_id} before subset construction. For fixed-budget SFT, we keep trajectories with at least 3 steps, \tqs coverage $\ge 4/7$, and tokenized length at most 32{,}768, leaving \ph{38{,}382} eligible trajectories. From this pool we construct matched Top-$K$, Bottom-$K$, and Random-$K$ subsets of 1{,}024 trajectories, controlling task coverage, repository coverage, and source proportions. Table~\ref{tab:subsetstats} summarizes the resulting subsets. Sampler details are in Appendix~\ref{app:subset}.

\paragraph{Training and evaluation.}
All SFT runs use Qwen3-Coder-30B-A3B-Instruct~\citep{yang2025qwen3} with LoRA~\citep{hu2022lora}. The training recipe is fixed across conditions: adapters are placed on the attention projections, loss is applied to \texttt{assistant} and \texttt{tool\_call} segments only, and every hyperparameter except the selected trajectories is held constant (Appendix~\ref{app:impl}). We evaluate pass@1 on \swev~\citep{openai2024verified} using \openhands v0.52.1~\citep{wang2025openhands}, matching the scaffold and version used to generate the training trajectories. We also recompute \tqs components on evaluation rollouts to diagnose process-level behavioral changes.

\begin{table*}[t]
\centering
\small
\setlength{\tabcolsep}{6pt}
\renewcommand{\arraystretch}{1.1}
\begin{tabular}{@{}llcccc>{\columncolor[gray]{0.97}}c c@{}}
\toprule
& & & \multicolumn{3}{c}{\textbf{pass@1 by difficulty (\%)}} & & \\
\cmidrule(lr){4-6}
\textbf{ID} & \textbf{Method} & \textbf{$|\mathcal{D}|$}
            & \textbf{Easy}  & \textbf{Medium} & \textbf{Hard}
            & \textbf{Overall} & \textbf{$\Delta$Base} \\
\midrule
\multicolumn{8}{@{}l}{\emph{No-SFT control}} \\
B0    & Qwen3-Coder-30B-A3B (base, no SFT)   & ---  & 36.6 & 19.2 & 2.2 & 24.4 & --- \\
\midrule
\multicolumn{8}{@{}l}{\emph{Subset baselines (matched 1{,}024 trajectories, identical SFT recipe)}} \\
B1    & Random-$K$                                 & 1{,}024 & 40.7 & 16.9 & 2.2 & 24.8 & $+$0.4 \\
B2    & Bottom-$K$ (lowest \tqs)                   & 1{,}024 & 33.5 & 16.9 & 0.0 & 21.8 & $-$2.6 \\
B3    & Shortest-$K$ (length-only)                 & 1{,}024 & 37.6 & 18.4 & 0.0 & 24.2 & $-$0.2 \\
B4    & DPI-only-$K$ (guardrail only)              & 1{,}024 & 37.6 & 19.9 & 2.2 & 25.2 & $+$0.8 \\
B5    & LLM-Judge-$K$ (GPT-4o rubric)              & 1{,}024 & \textbf{43.3} & \textbf{21.1} & \textbf{6.7} & \textbf{28.4} & \textbf{$+$4.0} \\
B6    & Top-LM-$K$ (length-matched to Random)      & 1{,}024 & 40.7 & 20.7 & 4.4 & 27.0 & $+$2.6 \\
\midrule
\rowcolor[gray]{0.92}
\textbf{Ours} & \textbf{Top-$K$ (highest \tqs)}    & 1{,}024 & 42.3 & 20.3 & \textbf{6.7} & 27.6 & $+$3.2 \\
\bottomrule
\end{tabular}
\caption{\textbf{Main results on \swev.} All SFT runs use the same Qwen3-Coder-30B-A3B-Instruct base, LoRA hyperparameters, and \openhands evaluation scaffold; only the selected training trajectories differ. Difficulty buckets follow \citet{openai2024verified}.}
\label{tab:main}
\end{table*}

\paragraph{Baselines.}
We compare against the untuned base model, Random-$K$, Bottom-$K$, Shortest-$K$ by token count, \DPI-only-$K$, GPT-4o LLM-Judge-$K$ under an AgentDiagnose-style~\citep{ou2025agentdiagnose} rubric, and Top-LM-$K$, which length-matches high-\tqs trajectories to the Random-$K$ length distribution. All baselines are drawn from the same eligible pool and trained with the same SFT recipe.

\section{Results}
\label{sec:results}

\subsection{Main Results}
\label{sec:main}

Table~\ref{tab:main} reports the main results. Under the same backbone, SFT recipe, and evaluation scaffold, the selected trajectories have a clear effect on downstream performance. Random-$K$ improves only slightly over the base model, from 24.4 to 24.8 pass@1. In contrast, \tqs-Top reaches 27.6, improving over Random-$K$ by 2.8 points and over the base model by 3.2 points. Bottom-$K$ moves in the opposite direction, dropping to 21.8. This ordering shows that trajectories with the same reward-1 outcome can differ substantially in their value as supervision.

The simpler baselines do not explain the \tqs gain. Shortest-$K$ reaches 24.2, below Random-$K$, so selecting short trajectories alone is not sufficient as a selection criterion. DPI-only-$K$ gives a small improvement over Random-$K$ but remains 2.4 points below \tqs-Top. The length-matched control, Top-LM-$K$, reaches 27.0 after matching the length distribution of Random-$K$. It retains most of the Top--Random gain, which indicates that length is correlated with trajectory quality but does not fully account for the effect of \tqs.

\begin{figure}[t]
\centering
\includegraphics[width=\linewidth]{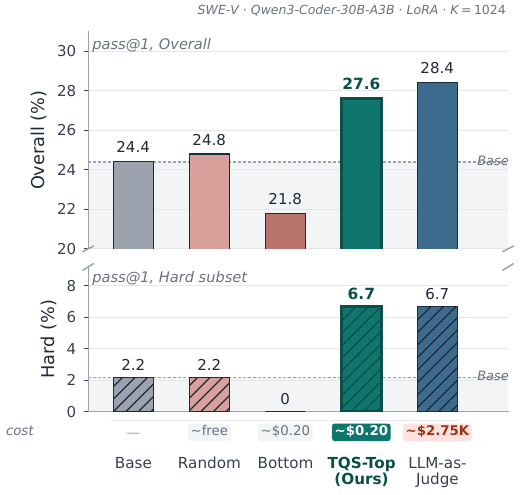}
\caption{\textbf{Main results on \swev.} Pass@1 on the full test set (top) and Hard subset (bottom). \tqs-Top recovers most of the LLM-as-Judge gain at much lower selection cost.}
\label{fig:main_result_vis}
\end{figure}

Among the non-\tqs selectors, LLM-Judge-$K$ is the strongest baseline at 28.4 pass@1. This is 0.8 points above \tqs-Top, showing that rubric-based GPT-4o judging remains competitive. Still, \tqs-Top recovers most of its improvement over Random-$K$ and matches LLM-Judge-$K$ on Hard tasks at 6.7. Figure~\ref{fig:main_result_vis} summarizes this trade-off between accuracy and selection cost.

Their selection costs differ more sharply. Under the prompt, token-count, and pricing assumptions used for the baseline, GPT-4o judging is estimated to cost \ph{$\sim\!\$2{,}750$} for the 38{,}382 eligible trajectories and \ph{$\sim\!\$18.6K$} for all 258{,}134 unique trajectories (Appendix~\ref{app:judge}). In comparison, the current \tqs implementation processes a 3{,}462-trajectory benchmark (median 53 steps) at 43 ms per trajectory with one worker on an AMD EPYC 7413. Extrapolating this measured rate gives about 3.1 CPU-hours for the unique corpus. With four workers, the measured throughput is 87.5 trajectories per second, corresponding to about 49 minutes for a full-corpus pass, or roughly \$0.1--\$0.2 in on-demand cloud CPU time at representative 4-vCPU rates (Appendix~\ref{app:impl}). These measurements make \tqs suitable for corpus-scale ranking, while semantic judging can be reserved for settings where its additional cost is acceptable.

The benefit of \tqs selection is larger on harder tasks. The Top--Random gap increases from 1.6 points on Easy to 3.4 on Medium and 4.5 on Hard. This pattern suggests that generic successful supervision may suffice for easier problems, while more difficult tasks are more sensitive to the structure and quality of the selected trajectories.

\subsection{Direct Validation among Successful Runs}
\label{sec:direct-validation}

\begin{table}[t]
\centering
\small
\setlength{\tabcolsep}{4.2pt}
\renewcommand{\arraystretch}{1.08}
\begin{tabular}{l c c}
\toprule
\textbf{Validation statistic} & \textbf{Human} & \textbf{Sonnet-5} \\
\midrule
High vs. Low agreement
& 85.2 (208/244)
& 81.1 (198/244) \\

Mid vs. Low agreement
& 74.2 (190/256)
& 77.0 (197/256) \\

High vs. Mid agreement
& 76.8 (172/224)
& 67.4 (151/224) \\

\midrule
Overall agreement
& \textbf{78.7} (570/724)
& 75.4 (546/724) \\

Pooled concordance
& 73.3 (631/861)
& \textbf{76.2} (656/861) \\

Full-order match
& \textbf{45.7} (137/300)
& 41.3 (124/300) \\

Spearman $\rho$
& \textbf{0.417}
& 0.402 \\
\bottomrule
\end{tabular}

\caption{Direct validation of the \tqs ordering among successful runs. Agreement (\%) uses order-consistent pairwise judgments; pooled concordance uses pointwise rubric scores with ties excluded.}
\label{tab:direct-validation}
\end{table}

We next test whether the \tqs ordering agrees with content-based judgments while holding both task and outcome fixed. For each of 300 tasks with at least three resolved runs, we construct high--low, mid--low, and high--mid pairs, yielding 900 comparisons. Five software-engineering students split the human evaluation, with each pair assigned to one student, while Claude Sonnet-5 evaluates the same pairs separately. Both evaluations are blinded to \tqs and its components. Each pair is presented in both orders; 176 pairs receive inconsistent verdicts and are treated as ties, leaving 724 order-consistent comparisons.

Table~\ref{tab:direct-validation} shows consistent agreement across the three validation views. On the 724 order-consistent pairs, human and Sonnet-5 judgments agree with the \tqs ordering on 78.7\% and 75.4\% of comparisons, respectively, with the strongest agreement on high--low pairs (85.2\% and 81.1\%). Pointwise rubric scores yield similar pooled concordance, at 73.3\% for humans and 76.2\% for Sonnet-5. More stringently, the complete high$>$mid$>$low ordering matches \tqs on 45.7\% and 41.3\% of tasks, well above the 16.7\% chance rate for three-item rankings. Together, these results show that \tqs captures a substantial portion of the quality differences recognized by content-based evaluators, while remaining deliberately narrower than a semantic judge. Appendix~\ref{app:direct-validation} reports the full statistics and examples of semantic distinctions not captured by \tqs.

\subsection{Ablation of \tqs Dimensions}
\label{sec:dimension-ablation}

\begin{table}[t]
\centering
\small
\setlength{\tabcolsep}{4.5pt}
\renewcommand{\arraystretch}{1.08}
\begin{tabular}{l c}
\toprule
\textbf{Selection signal} & \textbf{pass@1} \\
\midrule
Full \tqs-Top & \textbf{27.6} \\
w/o Information focus & 24.6 \\
w/o Intent--action alignment & 23.0 \\
w/o Policy dynamics & 25.8 \\
\midrule
Random-$K$ & 24.8 \\
Base (no SFT) & 24.4 \\
\bottomrule
\end{tabular}
\caption{Dimension ablation of \tqs. Each variant removes one structural dimension from data selection and retrains with the same SFT recipe. Random-$K$ and the untuned base model are shown for reference.}
\label{tab:dimension-ablation}
\end{table}

We next test whether \tqs's downstream benefit comes from one structural dimension or their combination. From the full scorer, we remove information focus, intent--action alignment, or policy dynamics from data selection, reconstruct the Top-$K$ subset, and retrain with the same SFT recipe.

Table~\ref{tab:dimension-ablation} shows that all three dimensions contribute to the downstream gain. Removing intent--action alignment produces the largest drop, from 27.6 to 23.0 pass@1, falling below both Random-$K$ and the untuned base model. Removing information focus reduces performance to 24.6, while removing policy dynamics gives 25.8. None of the ablated variants matches the full \tqs result, indicating that the gain is not carried by a single dimension. The particularly large drop without intent--action alignment suggests that selecting trajectories whose stated reasoning is consistently reflected in subsequent actions is especially important for effective SFT supervision.

\subsection{Behavioral Diagnosis on Rollouts}
\label{sec:analysis}

\begin{table}[t]
\centering
\small
\setlength{\tabcolsep}{4.5pt}
\renewcommand{\arraystretch}{1.08}
\begin{tabular}{l c @{\hspace{10pt}} c c c}
\toprule
& \multicolumn{1}{c}{\textbf{Training Data}}
& \multicolumn{3}{c}{\textbf{Fine-tuned models}} \\
\cmidrule(lr){2-2} \cmidrule(lr){3-5}
\textbf{Process metric} & \textbf{Top-$K$} & \textbf{Top} & \textbf{Rand} & \textbf{Bottom} \\
\midrule
Mean Steps & 34.2  & \textbf{50.66} & 57.64 & 63.46 \\
Mean \OEC  & 0.603 & \textbf{0.571} & 0.521 & 0.484 \\
Mean \PSN  & 0.753 & \textbf{0.570} & 0.548 & 0.568 \\
Mean \SCP  & 0.994 & 0.777 & \textbf{0.798} & 0.772 \\
Mean \PED  & 0.826 & 0.192 & \textbf{0.267} & 0.164 \\
Mean \DPI  & 0.996 & \textbf{0.931} & 0.923 & 0.906 \\
\bottomrule
\end{tabular}
\caption{Process metrics of the Top-$K$ training data and the fine-tuned models' own \swev rollouts.}
\label{tab:behavior}
\end{table}

Finally, we ask whether data selection is reflected in the models' own test-time behavior. We recompute the \tqs components on their \swev rollouts and report the means in Table~\ref{tab:behavior}. The Top model uses 50.66 steps on average, compared with 57.64 for Random and 63.46 for Bottom. It also has higher \OEC and \DPI than both models; its \PSN is higher than Random and close to Bottom. The pattern is not uniform across all components in these rollouts, since Random has slightly higher \SCP and \PED than Top. This suggests that selecting high-\tqs supervision leaves a measurable behavioral imprint on the fine-tuned policy rather than affecting only final task accuracy. At the same time, the mismatch on \SCP and \PED indicates that the model does not simply reproduce the component profile of its training data. The clearest rollout differences are shorter interaction, more focused observations, and fewer dirty patterns.


\section{Conclusion}
\label{sec:conclusion}

"Passing" is too coarse a quality signal for coding-agent SFT data. We introduced \tqs, a lightweight event-log scorer that evaluates successful trajectories through information focus, intent--action alignment, policy dynamics, and a guardrail for categorical failure patterns. Across our experiments, high-\tqs trajectories provide more effective supervision than simpler outcome-, length-, or pattern-based selection, while low-\tqs trajectories can even hurt downstream performance. At the same time, \tqs approaches the effectiveness of rubric-based GPT-4o judging at substantially lower selection cost, making repeated rescoring and corpus-scale data selection practical. These results suggest that successful trajectories are not equally valuable as supervision, and that trajectory structure offers a useful, scalable signal beyond task outcome alone.

\section*{Limitations}

Our experiments use \coderforge trajectories generated by a single teacher model under the \openhands scaffold. Although the components rely on general event-log fields, their parsers and thresholds may require recalibration for different action spaces, multilingual traces, or other step granularities. Transfer to GUI or multimodal agents would also require new definitions of scope and intent. Outside the current setting, the hand-set component weights may likewise require recalibration. \tqs also scores the structure of an observed process rather than verifying whether a passing patch addresses the underlying root cause.

Our SFT experiments focus on a fixed-budget setting with 1{,}024 selected trajectories per condition. We also restrict training trajectories to at most 32K tokens so that they fit within the 40{,}960-token context after template overhead. This excludes a long-trajectory tail that may contain useful recovery behavior. We train each condition once because of compute cost, so the reported pass@1 differences do not include multi-seed variance estimates.

Our downstream evaluation is limited in model scale. Preliminary experiments with a 7B backbone did not show stable selector differences, so we focus the reported SFT comparison on the 30B model. Larger backbones remain untested.

\section*{Acknowledgments}

We thank the anonymous reviewers for their constructive feedback. This work was supported by the Fundamental Research Funds for the Central Universities, Peking University. Xu Sun is the corresponding author of this paper.

\bibliography{custom}
 
\appendix

\section{Worked-Example Component Scores}
\label{app:worked-example}

Table~\ref{tab:worked-example} gives the component breakdown for the two \texttt{pygments} trajectories in Figure~\ref{fig:case_study}. The table uses the fixed weights defined in \S\ref{sec:components}.

\section{Direct Validation and Semantic Gaps}
\label{app:direct-validation}

\paragraph{Within-task protocol.}
We sample 300 tasks with at least three resolved runs: 181 from SWE-Smith, 85 from SWE-Rebench, and 34 from R2E-Gym. High, middle, and low \tqs tiers yield 900 pairs. Each pair is presented in both orders, giving 1,800 judgments in the human study and another 1,800 in the Claude Sonnet-5 study. Five software-engineering students split the human labeling, with one student assigned to each pair. Neither study sees \tqs or its components. The rubric covers problem comprehension, solution quality and minimality, exploration efficiency, error recovery, and verification rigor. Of the 900 pairs, 176 reverse under presentation order and are treated as ties. The main table reports the remaining 724 pairs, or 80.4\% coverage.

\begin{table}[t]
\centering
\small
\setlength{\tabcolsep}{5pt}
\begin{tabular}{@{}lccc@{}}
\toprule
\textbf{Component} & \textbf{Weight} & \textbf{Run 2} & \textbf{Run 4} \\
\midrule
\OEC & 0.30 & 0.7384 & 0.3120 \\
\IAC & 0.25 & 0.6667 & 0.5610 \\
\PSN & 0.10 & 0.8491 & 0.5708 \\
\TTE & 0.10 & 0.5718 & 0.4046 \\
\SCP & 0.05 & 0.9904 & 0.5280 \\
\PED & 0.10 & 1.0000 & 0.0000 \\
\DPI & 0.10 & 1.0000 & 0.6000 \\
\midrule
\textbf{TQS} & & \textbf{0.7798} & \textbf{0.4178} \\
\bottomrule
\end{tabular}
\caption{Component scores for the two resolved runs in Figure~\ref{fig:case_study}. Run 2 is the high-\tqs trajectory and Run 4 is the low-\tqs trajectory.}
\label{tab:worked-example}
\end{table}

\begin{table*}[t]
\centering
\scriptsize
\renewcommand{\arraystretch}{1.05}

\begin{tabular*}{\textwidth}{
@{\extracolsep{\fill}}
l
ccc
ccc
@{}
}
\toprule
& \multicolumn{3}{c}{\textbf{Human}}
& \multicolumn{3}{c}{\textbf{Claude Sonnet-5}} \\
\cmidrule(lr){2-4}
\cmidrule(lr){5-7}

\textbf{Contrast}
& \textbf{Agreement}
& \textbf{95\% CI}
& \textbf{$p$}
& \textbf{Agreement}
& \textbf{95\% CI}
& \textbf{$p$} \\
\midrule

High vs. Low
& 85.2\% (208/244)
& 80.3--89.1
& $1.34\!\times\!10^{-30}$
& 81.1\% (198/244)
& 75.8--85.6
& $1.18\!\times\!10^{-23}$ \\

Mid vs. Low
& 74.2\% (190/256)
& 68.5--79.2
& $4.27\!\times\!10^{-15}$
& 77.0\% (197/256)
& 71.4--81.7
& $1.51\!\times\!10^{-18}$ \\

High vs. Mid
& 76.8\% (172/224)
& 70.8--81.8
& $3.43\!\times\!10^{-16}$
& 67.4\% (151/224)
& 61.0--73.2
& $2.04\!\times\!10^{-7}$ \\

\midrule
\textbf{Overall}
& \textbf{78.7\% (570/724)}
& 75.6--81.6
& $5.94\!\times\!10^{-57}$
& \textbf{75.4\% (546/724)}
& 72.1--78.4
& $2.70\!\times\!10^{-44}$ \\

\bottomrule
\end{tabular*}

\caption{Agreement with the \tqs ordering among order-consistent pairs. Confidence intervals are 95\% Wilson intervals. The $p$-values use two-sided exact binomial tests against 0.5.}
\label{tab:direct-validation-full}
\end{table*}

The rubric scores also give a pointwise comparison that does not require the two presentations to produce the same discrete verdict. Table~\ref{tab:rubric-concordance} reports pairwise concordance after excluding equal-score ties. Across the three pair types, the human rubric agrees with \tqs on 631/861 pairs and Claude Sonnet-5 agrees on 656/861. The full high$>$mid$>$low ordering matches \tqs for 137/300 tasks under the human rubric and 124/300 tasks under Claude Sonnet-5, compared with $1/6$ of tasks under a random ordering.

\begin{table*}[t]
\centering
\scriptsize
\renewcommand{\arraystretch}{1.05}

\begin{tabular*}{\textwidth}{
@{\extracolsep{\fill}}
l
ccc
ccc
c
@{}
}
\toprule
& \multicolumn{3}{c}{\textbf{Human rubric}}
& \multicolumn{3}{c}{\textbf{Claude Sonnet-5 rubric}}
& \\
\cmidrule(lr){2-4}
\cmidrule(lr){5-7}

\textbf{Contrast}
& \textbf{Conc.}
& \textbf{95\% CI}
& \textbf{$p$}
& \textbf{Conc.}
& \textbf{95\% CI}
& \textbf{$p$}
& \textbf{Ties} \\
\midrule

High vs. Low
& 78.3\% (227/290)
& 73.2--82.6
& $6.52\!\times\!10^{-23}$
& 82.1\% (238/290)
& 77.2--86.1
& $1.35\!\times\!10^{-29}$
& 10 \\

Mid vs. Low
& 69.5\% (203/292)
& 64.0--74.5
& $2.11\!\times\!10^{-11}$
& 75.7\% (221/292)
& 70.5--80.3
& $4.39\!\times\!10^{-19}$
& 8 \\

High vs. Mid
& 72.0\% (201/279)
& 66.5--77.0
& $1.11\!\times\!10^{-13}$
& 70.6\% (197/279)
& 65.0--75.6
& $4.39\!\times\!10^{-12}$
& 21 \\

\midrule
\textbf{Pooled}
& \textbf{73.3\% (631/861)}
& 70.2--76.1
& $6.59\!\times\!10^{-44}$
& \textbf{76.2\% (656/861)}
& 73.2--78.9
& $1.05\!\times\!10^{-55}$
& 39 \\

\bottomrule
\end{tabular*}

\caption{Concordance between pointwise rubric scores and the \tqs tier ordering after excluding equal-score ties. Confidence intervals are 95\% Wilson intervals; $p$-values use two-sided exact binomial tests against 0.5.}
\label{tab:rubric-concordance}
\end{table*}

\paragraph{What semantic judging adds.}
We inspect disagreements in which Claude Sonnet-5 prefers the lower-\tqs trajectory. On \texttt{python\_\_mypy-10430}, the preferred run traces a timing-dependent crash to a flag missing from \texttt{VAR\_FLAGS} across serialization and fixes it in one line. The higher-\tqs run duplicates lookup logic without resolving the timing failure. On the \texttt{sqlfluff} \texttt{func\_pm\_op\_change} task, the preferred run corrects an indexing error, while the higher-\tqs run handles a secondary tuple-subtraction failure and adds bounds checks around the original bug.

Verification produces another type of disagreement. On \texttt{wfondrie\_\_mokapot-65}, both runs apply the same one-line fix. The preferred run builds a reproduction, executes the existing test, and adds edge-case tests. The higher-\tqs run stops after static inspection without confirming a passing test. These examples show information available to semantic judging that \tqs does not encode.

\paragraph{Score geometry and selection stability.}
\tqs is a scalar functional, $\tqs:\mathcal{T}\rightarrow[0,1]$, rather than a structural distance between trajectories. Its role is to induce the total preorder used for subset selection: $\tau_i \succeq \tau_j$ whenever $\tqs(\tau_i)\geq\tqs(\tau_j)$. Distinct trajectories may therefore be equivalent for selection if they receive the same score. Accordingly,
\[
d(\tau_i,\tau_j)
=
\left|\tqs(\tau_i)-\tqs(\tau_j)\right|
\]
is a pseudometric on $\mathcal T$, and becomes a metric after identifying trajectories with equal scores. This only describes separation in score space; nearby TQS values do not imply structural similarity between trajectories.

For subset selection, the relevant quantity is the score margin. Let $s(\tau)=\tqs(\tau)$ and let $\tilde{s}$ be a perturbed scoring function such that
\[
\sup_{\tau\in\mathcal T}
\left|\tilde{s}(\tau)-s(\tau)\right|
\leq \epsilon.
\]
Then, for any pair $\tau_i,\tau_j$,
\[
\tilde{s}(\tau_i)-\tilde{s}(\tau_j)
\geq
s(\tau_i)-s(\tau_j)-2\epsilon.
\]
Hence, any pair whose original score gap exceeds $2\epsilon$ preserves its ordering. For scores sorted as $s_{(1)}\geq\cdots\geq s_{(N)}$, define the Top-$K$ boundary margin as $\Delta_K=s_{(K)}-s_{(K+1)}$. If $\Delta_K>2\epsilon$, the Top-$K$ subset is unchanged. More generally, only trajectories close to the selection boundary can enter or leave the selected set under a bounded perturbation.

This also gives a simple robustness condition for the TQS weights. Writing
\[
\tqs_{\mathbf w}(\tau)
=
\sum_{m=1}^{M} w_m\phi_m(\tau),
\qquad
\phi_m(\tau)\in[0,1],
\]
a perturbation $\delta\mathbf w$ satisfies
\[
\left|
\tqs_{\mathbf w+\delta\mathbf w}(\tau)
-
\tqs_{\mathbf w}(\tau)
\right|
\leq
\|\delta\mathbf w\|_1.
\]
Therefore, $\Delta_K>2\|\delta\mathbf w\|_1$ is a sufficient worst-case condition for preserving the Top-$K$ subset. This bound is conservative, while the empirical Top-$K$ overlap under weight perturbations directly measures the local stability of the selected subset around the chosen TQS configuration.

\section{Implementation Details}
\label{app:impl}

\paragraph{Parser.}
Trajectories are stored as \openhands-format parquet files with the columns \texttt{trajectory\_id}, \texttt{instance\_id}, \texttt{repo}, \texttt{messages}, \texttt{tools}, \texttt{reward}, and \texttt{finish\_reason}. Our parser converts each row into the step tuple in \S\ref{sec:tqs}. Tool calls are recovered from the OpenAI-style \texttt{messages}/\texttt{tool\_calls} schema, and observations are attached to the corresponding tool step whenever the event log contains a returned tool message. File paths are extracted from tool arguments and normalized before scope-based measures are computed.

\paragraph{Scoring cost.}
The scorer is a deterministic CPU pipeline, and each component is linear in trajectory length. We benchmark the current implementation on an AMD EPYC 7413 using 3{,}462 mixed-source trajectories with a median length of 53 steps. A single worker processes this batch in 149.2 seconds, or 23.2 trajectories per second (43 ms per trajectory). At this rate, scoring all 258{,}134 unique trajectories requires about 3.1 CPU-hours. Four workers reach 87.5 trajectories per second, which corresponds to about 49 minutes for a full-corpus pass. Profiling attributes 58\% of the single-worker runtime to regular-expression error detection and 13\% to the character-level entropy calculation used by \OEC.

\paragraph{Loss masking.}
For SFT, ms-swift computes loss on assistant responses, including tool calls, while masking user messages and tool responses. Under the Qwen3-Coder agent template, tool calls are rendered in XML. For example:
\begin{quote}\small
\texttt{<tool\_call>}\\
\texttt{<function=execute\_bash>}\\
\texttt{\quad<parameter=command>}\\
\texttt{\quad pytest tests/}\\
\texttt{\quad</parameter>}\\
\texttt{</function>}\\
\texttt{</tool\_call>}
\end{quote}
This setup trains the model to imitate both natural-language reasoning and tool invocation, while excluding user messages and tool observations from the supervised loss. As a result, the loss focuses on the agent's reasoning and actions, while tool outputs serve only as conditioning context.

\paragraph{Error flags.}
The binary error flag $\varepsilon_t$ is set from tool observations using a small set of string and regular-expression patterns, including common execution failures, missing-file messages, traceback markers, failed test invocations, and tool-level error statuses. The flag is intentionally conservative: it is used to detect repeated failure patterns and post-error behavior, not to infer semantic correctness of the final patch.

\subsection{Subset Sampler}
\label{app:subset}

We construct fixed-budget training subsets from the eligible pool described in \S\ref{sec:components}. Each subset contains 1{,}024 trajectories. Top-$K$ and Bottom-$K$ are selected by descending and ascending \tqs, respectively, while Random-$K$ is sampled with a fixed seed. To keep the comparison focused on
trajectory quality, the sampler controls three sources of variation: task identity, repository coverage, and source provenance.

For Top-$K$ and Bottom-$K$, we first sort eligible trajectories by \tqs\ in the corresponding direction. The sampler then fills the subset greedily. In the first pass, it selects at most one trajectory per task. In the second pass, it prioritizes trajectories from repositories not yet represented in the subset. In the final pass, it fills any remaining slots in score order, subject to a hard per-task cap of 30 trajectories. This preserves \tqs\ ranking as the primary selection criterion while avoiding subsets dominated by a small number of tasks or repositories.

Because \coderforge\ combines trajectories from multiple data sources, we also stratify each subset by source. Target source proportions are computed from the eligible pool and applied to Top-$K$, Bottom-$K$, and Random-$K$. Random-$K$ uses the same eligibility filters and source proportions, but samples uniformly
within each source stratum rather than sorting by \tqs. All subset construction is performed after deduplication by \texttt{trajectory\_id}. Ties at selection boundaries are broken deterministically by \texttt{trajectory\_id}.

\section{Score and Component Distributions}
\label{app:dist}
\begin{figure}[t]
\centering
\includegraphics[width=0.92\columnwidth]{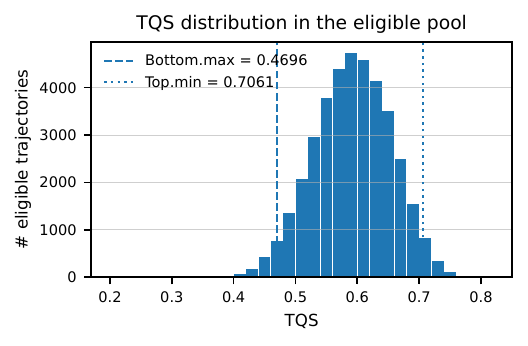}
\caption{\tqs distribution on the 38{,}382-trajectory eligible pool. The distribution peaks around 0.58--0.60. The Top-1024 threshold is 0.7061, and the Bottom-1024 threshold is 0.4696, leaving a gap of 0.2365 between the two selected subsets.}
\label{fig:score-dist}
\end{figure}

\begin{table}[t]
\centering
\small
\setlength{\tabcolsep}{5pt}
\renewcommand{\arraystretch}{1.08}
\begin{tabular}{@{}lccc@{}}
\toprule
\textbf{Component} & \textbf{Mean} & \textbf{Std.} & \textbf{Undefined} \\
\midrule
\OEC & 0.429 & 0.118 & 0.0\% \\
\PSN & 0.646 & 0.153 & 0.0\% \\
\IAC & 0.570 & 0.098 & 0.0\% \\
\TTE & 0.550 & 0.038 & 0.0\% \\
\SCP & 0.978 & 0.060 & 0.0\% \\
\PED & 0.580 & 0.232 & 0.7\% \\
\DPI & 0.941 & 0.127 & 0.0\% \\
\bottomrule
\end{tabular}
\caption{Marginal statistics of the seven retained \tqs components on the eligible pool. \SCP and \DPI are concentrated near 1, while \OEC, \PSN, \IAC, and \PED provide broader variation.}
\label{tab:component-summary}
\end{table}

Figure~\ref{fig:score-dist} shows the distribution of the final \tqs\ over the 38{,}382 eligible trajectories. The scores span a broad range rather than collapsing into a narrow band, providing sufficient resolution for fixed-budget ranking. The Top-1{,}024 cutoff is 0.7061, whereas the Bottom-1{,}024 cutoff is 0.4696, leaving a score gap of 0.2365 between the two selected regions. Random-$K$, by construction, remains centered around the eligible-pool distribution rather than either tail. This separation is important operationally: \tqs\ does not merely reorder trajectories with nearly identical scores, but identifies distinct high- and low-scoring regions from which fixed-size subsets can be constructed. The separation alone does not establish downstream quality, but provides the score resolution required for the selection experiments in \S\ref{sec:results}.

\begin{figure*}[t]
\centering
\begin{subfigure}{0.24\linewidth}
\includegraphics[width=\linewidth]{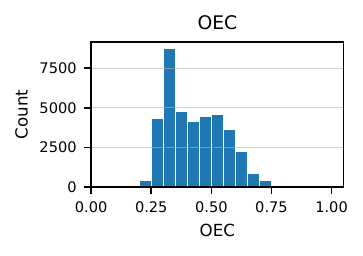}
\caption{\OEC}
\end{subfigure}
\begin{subfigure}{0.24\linewidth}
\includegraphics[width=\linewidth]{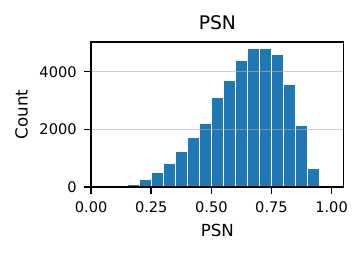}
\caption{\PSN}
\end{subfigure}
\begin{subfigure}{0.24\linewidth}
\includegraphics[width=\linewidth]{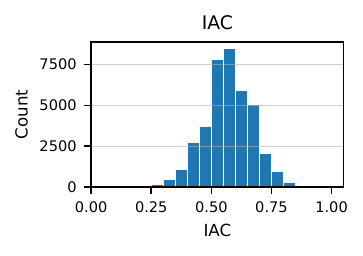}
\caption{\IAC}
\end{subfigure}
\begin{subfigure}{0.24\linewidth}
\includegraphics[width=\linewidth]{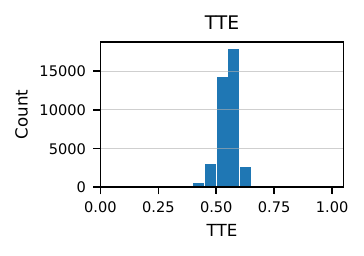}
\caption{\TTE}
\end{subfigure}
\begin{subfigure}{0.24\linewidth}
\includegraphics[width=\linewidth]{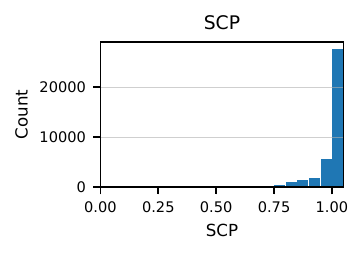}
\caption{\SCP}
\end{subfigure}
\begin{subfigure}{0.24\linewidth}
\includegraphics[width=\linewidth]{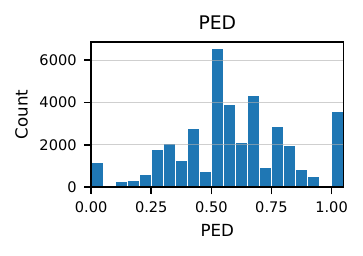}
\caption{\PED}
\end{subfigure}
\begin{subfigure}{0.24\linewidth}
\includegraphics[width=\linewidth]{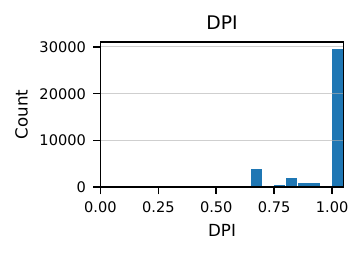}
\caption{\DPI}
\end{subfigure}
\caption{Marginal distributions of the seven retained \tqs components on the eligible pool. All components are normalized to $[0,1]$ before aggregation.}
\label{fig:component-dist}
\end{figure*}

\begin{figure}[t]
\centering
\includegraphics[width=0.92\columnwidth]{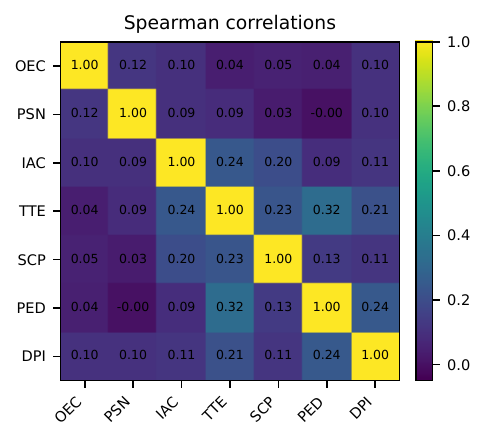}
\caption{Spearman correlation matrix among retained \tqs components on the eligible pool. The largest off-diagonal correlation is 0.320 between \TTE and \PED, indicating no severe redundancy among the retained components.}
\label{fig:component-corr}
\end{figure}

Table~\ref{tab:component-summary} and Figure~\ref{fig:component-dist} characterize the seven retained components individually. Their marginal distributions differ substantially, consistent with their different roles in structured convergence. \PED\ has the largest dispersion ($\sigma=0.232$), followed by \PSN\ ($\sigma=0.153$) and \OEC\ ($\sigma=0.118$), providing substantial variation across the eligible pool. \IAC\ also varies meaningfully around its mean of 0.570, whereas \TTE\ is more tightly concentrated ($\sigma=0.038$). In contrast, \SCP\ and \DPI\ have high means of 0.978 and 0.941, respectively, because most eligible trajectories already avoid their most severe failure patterns. Their value therefore lies primarily in identifying the smaller tail of runs with weak commitment or dirty interaction patterns rather than providing continuous discrimination across the entire pool.

The components also differ in coverage. Six are defined for every eligible trajectory, while \PED\ is undefined for only 0.7\% of runs because its error-recovery signal requires an applicable error transition. This small amount of missingness is handled by the renormalized aggregation in Eq.~\ref{eq:tqs}, rather than assigning an artificial value to trajectories for which the component cannot be evaluated. Overall, the heterogeneous marginals show that the final score combines both broadly varying signals such as \OEC, \PSN, and \PED\ and more selective guardrail-like signals such as \SCP\ and \DPI.

Figure~\ref{fig:component-corr} further examines whether these signals are redundant. Pairwise Spearman correlations remain modest across the retained components; the largest off-diagonal correlation is only 0.320, between \TTE\ and \PED. Thus, no component is well approximated by another through a simple monotonic relationship. This is consistent with the intended design of \tqs: information focus, intent--action alignment, policy dynamics, and categorical failure patterns contribute overlapping but distinct structural evidence. The correlation analysis by itself does not show that every component is necessary, but it rules out severe marginal redundancy; the downstream dimension ablations in \S\ref{sec:dimension-ablation} provide the complementary test of whether these signals matter for data selection.

\section{Failed Candidate Measures}
\label{app:failed}

We also considered several candidate measures that were motivated by the same structured-convergence view but failed the admission protocol in \S\ref{sec:validity}. Their definitions are included here to clarify the boundary of the final score.

\paragraph{HCR (Hypothesis--Commit Ratio).}
HCR characterizes how an agent allocates steps between exploration and execution. We define exploration as
$\mathcal{R}=\{\texttt{think},\texttt{read\_file},\texttt{search}\}$ and
execution as
$\mathcal{E}=\{\texttt{write\_file},\texttt{execute\_bash}\}$, and compute their ratio within the early, middle, and late thirds of the trajectory:
\[
R_g=
\frac{\#\{t\in G_g:a_t\in\mathcal R\}}
     {\#\{t\in G_g:a_t\in\mathcal E\}}.
\]
HCR assigns score 1 when $R_{\mathrm{early}}\ge R_{\mathrm{mid}}\ge R_{\mathrm{late}}$, favoring an
explore-then-commit pattern, and 0 otherwise. Although it passes the numerical thresholds (concordance 0.527, $|\rho|=0.014$), including HCR caused 99.9\% of the Top-1K subset to share the same binary pattern, so we exclude it under the behavioral-diversity criterion.

\paragraph{BUS (Backtracking Utility Score).}
BUS distinguishes productive redirection from unproductive wandering. A backtracking event is detected when the agent's current target-file set has no overlap with that of the preceding $k$ steps. The event is considered useful if, within the following $k$ steps, the search scope narrows or the
agent makes a substantive edit:
\[
s_{\mathrm{BUS}}
=
\frac{\#\text{ useful backtracking events}}
     {\#\text{ backtracking events}}.
\]
Higher BUS therefore indicates that changes in direction more often lead to focused progress. Its concordance is only 0.202, so we exclude it.

\paragraph{TGCR (Tool-Graph Cycle Ratio).}
TGCR captures structural redundancy in file navigation. We construct a directed graph whose nodes are files and whose edges represent successive file-access transitions, and use DFS coloring to identify back edges:
\[
s_{\mathrm{TGCR}}
=
\frac{\#\text{ back edges}}
     {\#\text{ total edges}}.
\]
A larger value indicates more cyclic revisitation, whereas a trajectory closer to a DAG reflects a cleaner access path. TGCR obtains concordance 0.479, close to chance, and is therefore excluded.

\paragraph{DCS-Jaccard (Decision Consistency Score).}
DCS measures local consistency between reasoning, action, and subsequent feedback. For each step, it combines intent--action alignment with the extent to which the resulting observation informs the next intent:
\[
\begin{aligned}
s_{\mathrm{DCS}}
&=
\frac{1}{2}
\Big[
\mathrm{sim}(intent_t,action_t) \\
&\qquad+
\mathrm{sim}(obs_t,intent_{t+1})
\Big].
\end{aligned}
\]
DCS-Jaccard instantiates $\mathrm{sim}(\cdot,\cdot)$ with Jaccard
similarity and averages the resulting scores over valid steps. Although intended to capture coherent local transitions, it has
$|\rho|=0.729$ with step count and is therefore rejected as strongly length-confounded.

\paragraph{DCS-file (file-alignment variant).}
DCS-file replaces lexical Jaccard overlap with file-level alignment when evaluating the same intent--action and observation--intent transitions. It therefore asks whether the files implied by the agent's reasoning agree with those actually acted on and whether observed files are reflected in the next plan. The resulting score obtains concordance 0.238, substantially below chance in the intended direction, and is excluded.

\section{LLM-Judge Baseline Rubric}
\label{app:judge}

The B5 baseline uses GPT-4o as a trajectory quality judge under an
AgentDiagnose-style rubric. The judge sees the issue description, the trajectory's reasoning and tool-call sequence, and the four rubric dimensions below. It does not see the reference patch or held-out test outcome. Each dimension is scored on a 1--5 Likert scale, and the trajectory score is the mean of the four dimensions. The top 1{,}024 trajectories by mean score, after the
same eligibility filters and subset constraints as Top-$K$, form the B5 training subset.

\paragraph{Planning.}
Does the agent decompose the issue into a coherent plan before acting, and does the plan track the issue's structure rather than the first failure mode that comes to mind?

\paragraph{Grounding.}
Are claims and edits grounded in observations that the trajectory has actually surfaced, or does the agent assert facts about the codebase without prior evidence?

\paragraph{Reflection.}
When an action fails, does the agent diagnose the failure before retrying, or does it loop on minor variations of the same action?

\paragraph{Efficiency.}
Does each step advance the trajectory toward a fix, or does the trajectory contain redundant exploration, repeated reads of already-viewed files, or stuck-process interrupts?

\paragraph{Cost estimate.}
On the \ph{38{,}382}-trajectory eligible pool, with approximately \ph{28K} input tokens and \ph{200} output tokens per call, GPT-4o pricing of \$2.50/M input tokens and \$10/M output tokens gives a per-pool cost of roughly \ph{$\sim\!\$2{,}750$}. Applying the same judge to all \ph{258{,}134} unique trajectories would cost roughly \ph{$\sim\!\$18.6K$}.

\section{Length-Controlled Pair-AUC}
\label{app:lengthauc}

Table~\ref{tab:lengthauc} gives a stricter length-confound check than the $|\rho|<0.25$ criterion in Table~\ref{tab:validity}. We restrict trajectories to the modal step-count band, $\text{total\_steps}\in[25,40]$, and ask whether retained components still separate Top from Bottom. The composite \tqs\ remains perfectly separating in this band. Among the largest components, \OEC, \IAC, and \PED\ remain above 0.96 pair-AUC, while \PSN\ drops to 0.839. This drop is diagnostic: scope narrowing is partly entangled with trajectory length, but the composite score is not driven by this single component.

\begin{table}[t]
\centering
\small
\setlength{\tabcolsep}{5pt}
\begin{tabular}{l c}
\toprule
\textbf{Score / Component} & \textbf{Pair-AUC} \\
\midrule
\tqs composite & 1.000 \\
\OEC & 0.996 \\
\IAC & 0.963 \\
\PED & 0.961 \\
\PSN & 0.839 \\
\bottomrule
\end{tabular}
\caption{Length-controlled pair-AUC in the modal step-count band
$\text{total\_steps}\in[25,40]$.}
\label{tab:lengthauc}
\end{table}

\begin{table}[t]
\centering
\small
\setlength{\tabcolsep}{5pt}
\renewcommand{\arraystretch}{1.08}
\begin{tabular}{@{}lrrc@{}}
\toprule
\textbf{Source} & \textbf{Passing} & \textbf{Eligible} & \textbf{Retention} \\
\midrule
SWE-Smith & 89{,}501 & 23{,}569 & 26.3\% \\
SWE-Rebench & 44{,}739 & 7{,}821 & 17.5\% \\
R2E-Gym & 20{,}904 & 6{,}992 & 33.4\% \\
\midrule
All & 155{,}144 & 38{,}382 & 24.7\% \\
\bottomrule
\end{tabular}
\caption{Retention after applying the fixed-budget SFT eligibility filters to passing trajectories. All passing trajectories satisfy the step-count and coverage filters; the main reduction comes from the tokenized-length cap and subsequent deduplication by \texttt{trajectory\_id}.}
\label{tab:cutoff}
\end{table}

\section{Tokenized-Length Cutoff}
\label{app:cutoff}

The 32{,}768-token cap is a packing-budget choice for SFT. It fits inside the 40{,}960-token context window while leaving room for chat-template and assistant-tag tokens. Of the \ph{155{,}144} reward$=1$ trajectories, \ph{38{,}382} survive the eligibility filters and deduplication, giving a retention rate of \ph{24.7\%}. The discarded tail is dominated by long trajectories, especially from SWE-Smith.

The cap should be interpreted as part of the fixed-budget SFT setting. Longer trajectories may contain useful recovery behavior, but they require longer-context or sequence-parallel training to imitate faithfully. We leave this setting to future work.

\section{Sensitivity to Component Weights}
\label{app:weights}

The main experiments use the fixed, hand-set weights defined in \S\ref{sec:components}. They allocate 0.40 to information focus, 0.25 to intent--action alignment, 0.25 to policy dynamics, and 0.10 to the \DPI guardrail. No downstream pass@1 result was used to tune them. We test local selection stability by perturbing the two largest individual weights, $w_{\OEC}$ and $w_{\IAC}$, and the guardrail weight $w_{\DPI}$. For each setting, we rebuild the Top-$K$ subset and report its overlap with the main selection. This experiment measures how much the selected set changes near the chosen weights.

\begin{table}[t]
\centering
\small
\setlength{\tabcolsep}{6pt}
\renewcommand{\arraystretch}{1.08}
\begin{tabular}{@{}lc@{}}
\toprule
\textbf{Weight setting} & \textbf{Top-1024 overlap} \\
\midrule
Main weights & 100.0\% \\
$w_{\OEC}=0.25$ & 93.1\% \\
$w_{\OEC}=0.35$ & 93.6\% \\
$w_{\IAC}=0.20$ & 93.1\% \\
$w_{\IAC}=0.30$ & 94.1\% \\
$w_{\DPI}=0.05$ & 98.3\% \\
$w_{\DPI}=0.15$ & 99.6\% \\
\bottomrule
\end{tabular}
\caption{Weight sensitivity of \tqs selection on the eligible pool. We perturb one weight at a time while keeping the other weights fixed, then measure overlap between the resulting Top-1024 set and the main Top-1024 set.}
\label{tab:weights}
\end{table}

\begin{figure}[t]
\centering
\includegraphics[width=0.92\linewidth]{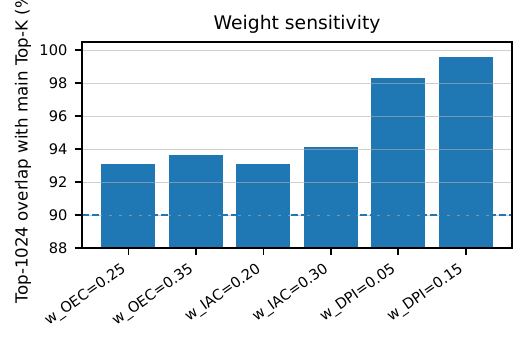}
\caption{Weight sensitivity of Top-1024 selection under single-weight perturbations.}
\label{fig:weight-sensitivity}
\end{figure}

Figure~\ref{fig:weight-sensitivity} visualizes the same result. OEC and IAC perturbations retain 93--94\% overlap with the main Top-1024 set, while DPI perturbations retain over 98\% overlap. These results indicate that the selected subset is locally stable to moderate changes in the hand-set weights. This analysis does not establish that the chosen weights are globally optimal, but it shows that small perturbations around them leave most of the selected training set unchanged.

\end{document}